\documentclass[sigconf]{acmart}

\setcopyright{none}
\usepackage[ruled,vlined]{algorithm2e} 

\SetAlFnt{\small}
\SetAlCapFnt{\small}
\SetAlCapNameFnt{\small}
\SetAlCapHSkip{0pt}

\acmJournal{TOG}
\usepackage{enumitem}
\usepackage{booktabs}
\usepackage{multirow}
\usepackage{graphicx}
\usepackage{caption}
\usepackage{pifont}

\definecolor{lightred}{RGB}{255,210,210}
\definecolor{lightorange}{RGB}{255,230,180}
\definecolor{lightyellow}{RGB}{255,248,180}

\copyrightyear{2026}
\acmYear{2026}
\setcopyright{cc}
\setcctype{by}
\acmConference[SA Posters '26]{SIGGRAPH Asia 2026 Posters}{December 01--04, 2026}{Kuala Lumpur, Malaysia}
\acmBooktitle{SIGGRAPH Asia 2026 Posters (SA Posters '26), December 01--04, 2026, Kuala Lumpur, Malaysia}
\acmDOI{10.1145/3829333.3847914}
\acmISBN{979-8-4007-2839-6/2026/12}

\begin{document}

\title{Geometry beneath the Waves: Dense Priors for Sparse-View Underwater 3D Gaussian Splatting}

\author{Harvey Caldeira}
\orcid{0009-0001-2484-9653}
\affiliation{\institution{University of Bristol}
\city{Bristol}
\country{United Kingdom}}
\email{jv21451@bristol.ac.uk}

\author{Haoran Wang}
\orcid{0009-0001-0388-445X}
\affiliation{\institution{University of Bristol}
\city{Bristol}
\country{United Kingdom}}
\email{yp22378@bristol.ac.uk}

\author{Guoxi Huang}
\orcid{0000-0002-8481-0232}
\affiliation{
\institution{University of Bristol}
\city{Bristol}
\country{United Kingdom}}
\email{guoxi.huang@bristol.ac.uk}

\author{Shaoyu Cai}
\orcid{0000-0001-8808-3442}
\affiliation{\institution{National University of Singapore}
\city{Singapore}
\country{Singapore}}
\email{shaoyucai@nus.edu.sg}

\author{Rachel Fu}
\orcid{0009-0006-7378-7029}
\affiliation{\institution{Oceanx}
\city{New York}
\country{USA}}
\email{rachel.fu@oceanx.org}

\author{Nantheera Anantrasirichai}
\orcid{0000-0002-2122-5781}
\affiliation{\institution{University of Bristol}
\city{Bristol}
\country{United Kingdom}}
\email{N.Anantrasirichai@bristol.ac.uk}

%

\renewcommand{\shortauthors}{Caldeira et al.}




\begin{teaserfigure}
  \includegraphics[width=0.98\textwidth]{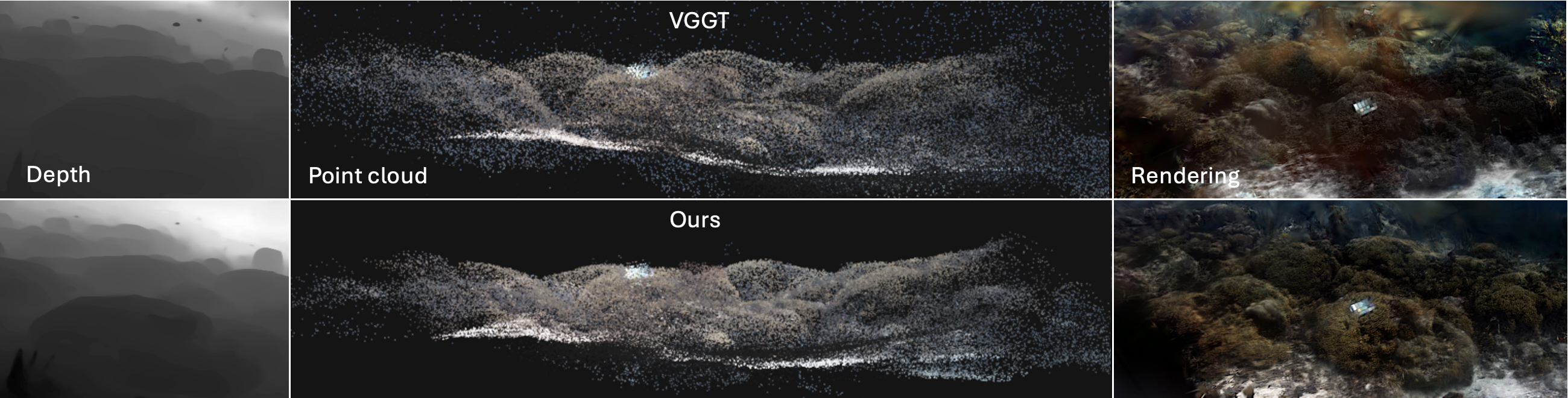}
  \caption{Qualitative comparison between VGGT (top) and our method (bottom) on the \emph{Curasao} scene. From left to right: rendered depth, reconstructed point cloud, and RGB rendering.}
  \Description{Underwater Geometry in a Forward Pass: Sparse-View 3D Reconstruction} \vspace{4mm}
  \label{fig:teaser}
\end{teaserfigure}


\maketitle

\section{Introduction}
Underwater 3D reconstruction supports applications ranging from marine ecosystem monitoring and subsea inspection to underwater archaeology, education, and immersive visualisation. 3D Gaussian Splatting (3DGS)~\cite{kerbl20233d} has made real-time photorealistic novel-view rendering practical, while underwater variants such as UW-GS~\cite{wang2025uw} and RUSplatting~\cite{jiang2025rusplatting} incorporate physically based image-formation models to separate medium effects from scene radiance. Their reconstruction quality, however, remains fundamentally limited by the geometry used for initialisation.
Water directly undermines this geometry. Wavelength-dependent absorption, backscatter, and suspended particulates reduce contrast and introduce high-frequency artefacts, degrading the feature correspondences required by classical Structure-from-Motion pipelines such as COLMAP~\cite{schonberger2016structure}. These failures become particularly severe under turbid conditions and sparse camera trajectories, where insufficient overlap further weakens geometric estimation.

Feed-forward geometry foundation models such as VGGT~\cite{wang2025vggt} offer an appealing alternative by regressing camera poses, dense depth, and point maps from unposed images in a single pass, without explicit feature matching. Applied zero-shot underwater, however, they exhibit two characteristic failure modes. First, \emph{volumetric water-column noise} arises when scattering and suspended particles are interpreted as solid geometry, filling open water with spurious point clusters. Second, \emph{high-frequency surface dispersion} causes predicted surfaces to fragment or jitter instead of converging to coherent boundaries. Although 3DGS initialised from these predictions often recovers the global scene structure, it also inherits both forms of geometric noise.

View sparsity introduces a second, complementary bottleneck. RUSplatting expands its feature tracks using temporal frame interpolation, but the resulting images are synthesised purely in 2D and are not constrained by the underlying scene geometry. When processed by a feed-forward reconstruction backbone, these photometrically and geometrically inconsistent frames can produce severe spatial artefacts. We address both limitations by adapting feed-forward geometry to the underwater domain and replacing image-space interpolation with geometry-guided view synthesis.

\section{Our approach}

\subsection{Domain-Adapted Feed-forward Geometry}

Reliable underwater geometric supervision is scarce: paired data are limited, while pseudo-labels from real images inherit the same scattering, absorption, and backscatter artefacts the model must overcome. Post-hoc alignment methods, including depth inversion, foreground scaling, histogram matching, and affine fitting, also fail to generalise across water types because the depth signals exhibit non-linear volumetric distortions that no global transformation can reconcile. We therefore reverse the supervision strategy: instead of adapting noisy depth targets, we preserve clean geometry and degrade the corresponding images. Our training set contains 22,000 images, comprising 1,000 Hypersim~\cite{roberts2021hypersim} scenes, 1,000 TartanAir~\cite{wang2020tartanair} images, and 20,000 unposed SA-1B~\cite{kirillov2023sam} images. We use metric ground-truth depth where available and Depth Anything 3~\cite{lin2026depth} pseudo-depth for SA-1B. The RGB images are then transformed using SyreaNet's~\cite{wen2023syreanet} physically guided model of wavelength-dependent absorption, scattering, and backscatter. By independently sampling attenuation coefficients, backscatter fractions, and transmission parameters for each frame, we expose the encoder to a broad range of turbidity levels and colour casts while retaining geometrically consistent supervision.

We adapt VGGT using the Fin3R~\cite{ren2025fin3r} teacher--student framework. A frozen teacher processes the clean image, while the student receives its synthetically degraded underwater counterpart. Training minimises
$\mathcal{L} = \lVert y_s - y_t \rVert^2$
encouraging the student to reproduce the teacher’s geometric predictions despite underwater appearance distortions. To preserve VGGT’s pretrained reconstruction capability, we insert LoRA~\cite{hu2022lora} adapters only into its visual aggregator encoder, while keeping all downstream prediction heads frozen. For each pretrained weight matrix $\mathbf{W}$, LoRA learns a low-rank update $\mathbf{W}' = \mathbf{W} + \mathbf{AB}$, with trainable matrices $\mathbf{A}$ and $\mathbf{B}$. The student therefore learns to encode degraded underwater imagery into the feature space expected by the original model, while preserving its pretrained geometric priors and metric-scale predictions.

\subsection{Geometry-Guided Radiance Field Rendering}

\begin{figure}[t]
  \centering
  \includegraphics[width=\linewidth]{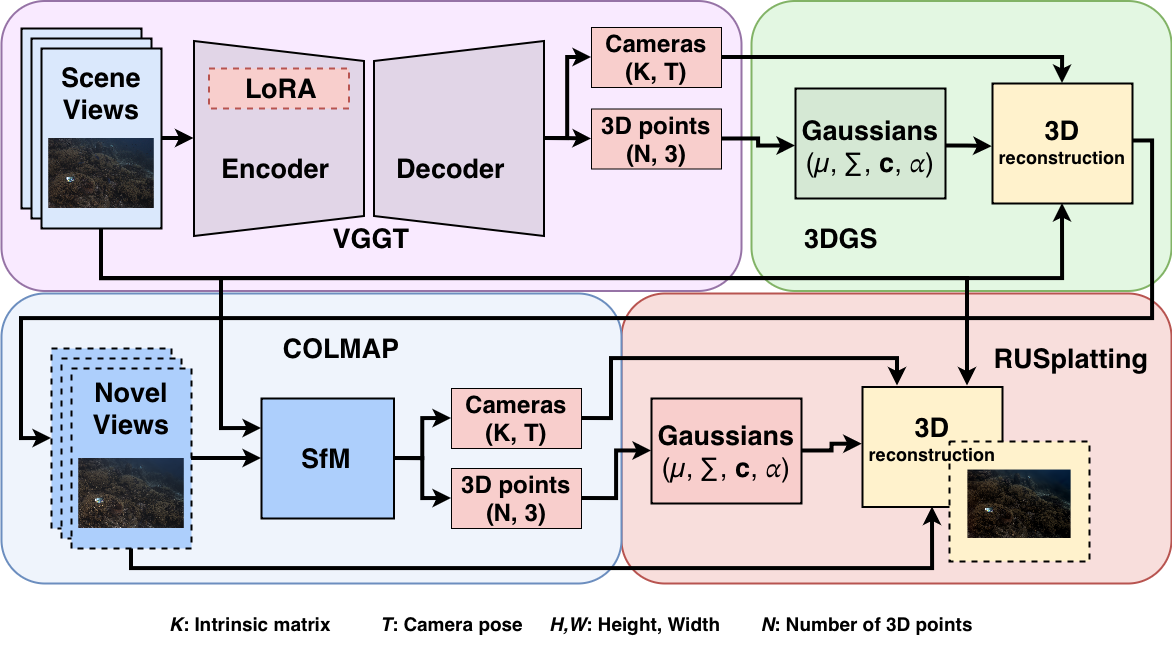} 
  \caption{Pipeline overview. LoRA-adapted VGGT predicts dense geometry, a 3DGS proxy renders geometry-guided pseudo-views, COLMAP expands tracks, and RUSplatting performs final optimisation.}
  \label{fig:pipeline} 
\end{figure}

Rather than densifying feature tracks through image-space interpolation, we generate additional observations from explicit 3D geometry (Fig.~\ref{fig:pipeline}). The fine-tuned VGGT first predicts a dense, structurally coherent point cloud from the unposed input views. This geometry initialises an intermediate 3DGS proxy, from which we render pseudo-views along the estimated camera trajectory. Each rendered view is paired with a pseudo depth map.

We then run COLMAP jointly over the captured and rendered views, increasing camera overlap and extending feature tracks across wider baselines. The resulting track graph and depth priors initialise RUSplatting, which performs the final radiance optimisation and underwater colour restoration. Unlike 2D flow-based interpolation, each pseudo-view is obtained by projecting an explicit 3D representation. The additional correspondences are therefore anchored to the estimated scene structure, providing dense geometric constraints for the final reconstruction.

\begin{table}[t]
  \caption{Performance comparison. Red, orange, and yellow indicate the best, second-, and third-best results, respectively.} 
  \label{tab:results}
  \resizebox{\columnwidth}{!}{%
  \begin{tabular}{@{}lccccccc@{}}
    \toprule
    & & \multicolumn{3}{c}{SeaThru-NeRF}
    & \multicolumn{3}{c}{Submerged3D} \\
    \cmidrule(lr){3-5}\cmidrule(l){6-8}
    Initialisation $\rightarrow$ Renderer
      & Iter.
      & PSNR$\uparrow$ & SSIM$\uparrow$ & LPIPS$\downarrow$
      & PSNR$\uparrow$ & SSIM$\uparrow$ & LPIPS$\downarrow$ \\
    \midrule
    COLMAP $\rightarrow$ 3DGS
      & 30k
      & \colorbox{lightorange}{25.68}
      & \colorbox{lightorange}{0.7735}
      & \colorbox{lightred}{0.1965}
      & \colorbox{lightorange}{23.48}
      & \colorbox{lightred}{0.7385}
      & \colorbox{lightorange}{0.3333} \\

    VGGT $\rightarrow$ 3DGS
      & 30k
      & 20.62
      & 0.5380
      & 0.3066
      & 19.38
      & 0.5001
      & 0.4915 \\

    FT-VGGT $\rightarrow$ 3DGS
      & 30k
      & 22.01
      & 0.6240
      & \colorbox{lightyellow}{0.2726}
      & 19.58
      & 0.4998
      & 0.4568 \\

    \midrule
    COLMAP $\rightarrow$ RUSplatting
      & 15k
      & \colorbox{lightyellow}{24.37}
      & \colorbox{lightyellow}{0.7611}
      & 0.3103
      & \colorbox{lightyellow}{22.97}
      & \colorbox{lightorange}{0.7114}
      & \colorbox{lightyellow}{0.3514} \\

    VGGT $\rightarrow$ RUSplatting
      & 15k
      & 19.76
      & 0.4824
      & 0.4463
      & 20.41
      & 0.5177
      & 0.4403 \\

    FT-VGGT $\rightarrow$ RUSplatting
      & 15k
      & 22.41
      & 0.6165
      & 0.3567
      & 19.34
      & 0.4857
      & 0.4597 \\

    \midrule
    \textbf{Ours (geometry-guided)}
      & 15k
      & \colorbox{lightred}{27.11}
      & \colorbox{lightred}{0.8634}
      & \colorbox{lightorange}{0.2188}
      & \colorbox{lightred}{23.49}
      & \colorbox{lightyellow}{0.7045}
      & \colorbox{lightred}{0.3322} \\
    \bottomrule
  \end{tabular}
  } 
\end{table}

\section{Experiments and Findings}

We fine-tune the LoRA adapters for 20 epochs using mixed precision on eight NVIDIA GH200 superchips, with a per-GPU batch size of 18 and four-step gradient accumulation, yielding an effective batch size of 576. The validation loss on 2,200 held-out synthetic images decreases steadily before plateauing, indicating stable convergence without evident overfitting.
We evaluate on SeaThru-NeRF~\cite{levy2023seathrunerf} and Submerged3D~\cite{jiang2025rusplatting}, each containing four scenes with approximately 20 views. Images are processed at 720p, with every eighth frame held out. Following the original settings, we use 30k iterations for 3DGS and 15k for RUSplatting, render one pseudo-view between consecutive frames, and retain up to $10^6$ VGGT points after applying a 30\% confidence threshold.

Table~\ref{tab:results} reports dataset averages. On the SeaThru-NeRF scenes, our pipeline improves COLMAP-initialised RUSplatting from 24.37 to 27.11 dB PSNR, with gains of 0.102 SSIM and 0.092 LPIPS. It also improves by up to 7.35 dB over raw feed-forward initialisation. On Submerged3D, our method achieves the best average PSNR of 23.49,dB and LPIPS of 0.3322, although gains are smaller due to greater lighting and colour variation.
VGGT adaptation also improves the geometric prior. On SeaThru-NeRF, PSNR rises from 20.62 to 22.01 dB with 3DGS and from 19.76 to 22.41 dB with RUSplatting, while LPIPS decreases from 0.4463 to 0.3567. Qualitatively, adaptation suppresses floating water-column points and recovers more coherent depth as shown in Fig.~\ref{fig:teaser}.

Two limitations remain: i) COLMAP fails on IUI3-RedSea dataset because scattering artefacts disrupt SIFT matching; ii) low-amplitude prediction noise persists after density control under sparse capture. These limitations motivate future work on geometry-aware pruning, semantic water-column masking, and iterative view synthesis to investigate whether refinement yields cumulative gains.

\bibliographystyle{ACM-Reference-Format}
\bibliography{main}

\end{document}